\documentclass{article}

\usepackage{cite}
\usepackage{graphicx}
\usepackage{geometry}
\usepackage{subfig}
\usepackage{booktabs}
\usepackage{amsfonts}
\usepackage{amsmath}
\usepackage{float}

\graphicspath{ {images/} }
\begin{document}

\vspace{1in}

\title{Lattice Map Spiking Neural Networks (LM-SNNs)\\ for Clustering and Classifying Image Data\\
~}

\author{Hananel Hazan$^{1}$, Daniel J. Saunders$^{1}$, Darpan T. Sanghavi$^{1}$, \\Hava Siegelmann$^{1}$, and Robert Kozma$^{1,2}$\\
~\\
$^{1}$ Biologically-Inspired Neural $\&$ Dynamical Systems Laboratory (BINDS)\\
University of Massachusetts Amherst, Amherst, MA01003, USA\\
~\\
$^{2}$ Center for Large-Scale Intelligent Optimization $\&$ Networks (CLION)\\
University of Memphis, TN38152, USA
}

\maketitle

\begin{abstract}

\textit{Spiking neural networks} (SNNs) with a lattice architecture are introduced in this work, combining several desirable properties of SNNs and \textit{self-organized maps} (SOMs). Networks are trained with biologically motivated, unsupervised learning rules to obtain a self-organized grid of filters via cooperative and competitive excitatory-inhibitory interactions. Several inhibition strategies are developed and tested, such as (i) \textit{incrementally increasing} inhibition level over the course of network training, and (ii) switching the inhibition level from low to high (\textit{two-level}) after an initial training segment. During the \textit{labeling} phase, the spiking activity generated by data with known labels is used to assign neurons to categories of data, which are then used to evaluate the network's classification ability on a held-out set of test data. Several biologically plausible evaluation rules are proposed and compared, including a population-level confidence rating, and an $n$-gram inspired method. The effectiveness of the proposed self-organized learning mechanism is tested using the MNIST benchmark dataset, as well as using images produced by playing the Atari Breakout game.

\end{abstract}
\vspace{0.1in}
\begin{center}
Original Manuscript Submitted: October 30, 2018\\
Revised: May 28, 2019\\
~\\
Special Issue: "Cognition and Neurocomputation" of\\
\it{Annals of Mathematics and Artificial Intelligence}
\end{center}

\newpage
\section{Introduction}
\label{intro}

Today's dominant \textit{artificial intelligence} (AI) approach uses \textit{deep neural networks} (DNNs), which are based on \textit{global} gradient-descent learning algorithms \cite{werbos74, pdp86}, wherein a loss function is defined and all DNN parameters are updated using approximate derivatives to minimize it. The success of this approach is based on employing massive amounts of data to train the DNNs, which requires significant computational resources \cite{y._lecun_deep_2015} provided by the exponentially increasing power of prolific supercomputing facilities world-wide. In the case of certain practical problems, however, one may not have a sufficiently large dataset to adequately cover the problem space, or one may need to make decisions quickly without waiting for an expensive training process. 

There are several proposed approaches to overcome the computational constraints of \textit{deep learning} (DL), one of which is based on using \textit{local} learning rules invoking neuro-biologically motivated spike-timing-dependent plasticity (STDP) \cite{bi_synaptic_1998}. Such learning rules represent a trade-off between the inevitably decreased classification performance due to the unsupervised nature of their operation, and the advantage provided by the distributed nature of the parameter updates. Namely, local learning rules do not require an end-to-end differentiable network structure, and no time is wasted on synchronous parameter updates through forward and backward passes widely used in back-propagation algorithms. Substituting global learning methods with local learning rules provides a solution to the computational bottleneck of deep learning, striking a balance between possibly increased learning speed and lesser memory requirements at the cost of reduced performance on machine learning (ML) tasks. 

In order to modify the synaptic parameters of spiking neural networks, one may record \textit{spike traces} and neuron \textit{membrane potentials}. The former captures a short-term trace of a neuron's spiking activity, while the latter is used in the simulation of the spiking neuron as part of its charging and action potential emission process. These two \textit{state variables} are all that is required to implement a robust set of correlation-based learning rules for synaptic strengths between spiking neurons; e.g., Hebbian learning \cite{hebb-organization-of-behavior-1949} and STDP \cite{10.3389/fnsyn.2012.00002}. Certain SNNs have the potential to be more energy efficient than their deep learning counterparts \cite{Merolla668, Cao2015}, as DNNs must cache the results of their layer-wise computation (produced during the \textit{forward pass}) to be used in the computation of gradients during the \textit{backward pass}. The memory requirement is proportional to the number and size of layers utilized by the network, and the size of the input data. SNNs which implement local learning rules are free from the forward-backward computational paradigm, and for that reason, may be beneficial in terms of time and memory complexity.

\subsection{Related Work}
\subsubsection{Spiking Neural Networks}
There is an extensive literature on spiking neural networks for various applications \cite{w._gerstner_spiking_2002}, but it was not until recently that SNNs demonstrated their feasibility to solve complex machine learning problems. Depending on the requirements of SNN simulations, various packages provide different levels of biological realism. For example, the NEST \cite{Gewaltig:NEST}, and BRIAN \cite{d._f._m._goodman_brian_2009} libraries are rather faithful to fine-grained neuro-biological details, while frameworks like CARLsim \cite{CARLsim} and Nengo \cite{bekolay2014} are more focused on engineering applications and efficiency, with less attention to biological constraints. 

The work by Diehl \& Cook \cite{p._u._diehl_unsupervised_2015} uses a simple three-layer spiking neural network to classify the MNIST handwritten digits by learning synaptic weights in an unsupervised fashion with several different STDP rules.  A number of other previous projects use STDP to train SNNs to classify image data \cite{liu_fast_2017, kheradpisheh_stdp-based_2016}. The former uses Gabor filters as a pre-processing input step to detect simple features, which are then used as input to their network \cite{liu_fast_2017}. They use a rank-order encoding scheme on the input spikes, which are processed through the network, in which the winner-take-all classification strategy is used on its output activity. The latter method comprises a difference of Gaussian pre-processing step followed by SNN with a series of convolutional and pooling layers, whose output is used to train a linear SVM for classification \cite{kheradpisheh_stdp-based_2016}. Other systems use spiking neurons, but require training with a supervisory signal; e.g., \cite{diehl_fast-classifying_2015}, in which a deep neural network is first trained using back-propagation and later transferred to a similar SNN without much loss in performance. More comprehensive artificial neural network (ANN) to SNN conversion methods are presented in \cite{10.3389/fnins.2017.00682, 2016arXiv161204052R}. Still others simply apply various approximations of the back-propagation algorithm directly to the spiking networks itself \cite{2017arXiv170604698H, 10.3389/fnins.2016.00508, Booij:2005:GDR:1195930.1710915, 1379954}.

\subsubsection{Self-Organizing Properties with Spiking Neural Networks}
There is a great potential to extend the work of \cite{p._u._diehl_unsupervised_2015,sala1998self} and \cite{saundersetal18} by combining STDP rules and competitive inhibitory interactions with a mechanism inspired by the self-organizing map (SOM) algorithm \cite{Kohonen1982} and the properties of the The adaptive resonance theory (ART) model \cite{Grossberg88}. SOMs are able to cluster an unlabeled dataset in an unsupervised manner. The topological arrangement created by the SOM algorithm forms specialized clusters that often correspond well with the categories that exist in the dataset. One caveat of the SOM training algorithm is that the size of the training set must be known in advance in order to adjust the learning rate accordingly. 

This clustering property is reminiscent of specialized areas in the primate cortex, wherein groups of neurons self-organize themselves based on the similarity of their role in functionality in parts of the primate body \cite{doi:10.1162/neco.1995.7.3.425}. An experimental study of Martinotti and basket cells provides an \textit{in silico} simulation of the self-organizing (clustering) ability of the cortex \cite{Amir_Noam_Hava}.

Early work use spiking neurons \cite{sala1998self} with similar topology used by \cite{Kohonen1982} for the benefit of self-organizing properties. The authors trained their network using a modified Hebbian learning with correlation function on a giving input. The algorithm showed selective behavioral activity and developed a receptive fields as expected from self-organized network. Unlike Kohonen SOM, this algorithm require to be initialize with unique random weights that is relative to the number of neurons in the receptive field. Moreover, it require tuning the strength of the learning magnitude to be proportional to the training size in order to reach a certain of conversion. Thus its not fit for training on the fly, where the size and the content of the dataset is unknown.  

Interesting work on self-organization in SNNs has been accomplished in \cite{6636061}, in which the authors analyze SOM formation in a network of integrate-and-fire (IF) neurons on both synthetic data and a cancer dataset. In \cite{YUSOB201357}, a SNN is first trained to form a SOM of a cancer dataset, after which it is used to classify the data. 

\subsection{Outline of the present work}

In this work, we start by employing the network architecture and code of Diehl $\&$ Cook \cite{p._u._diehl_unsupervised_2015}, which makes use of the BRIAN simulator. We introduce modifications to the original Diehl $\&$ Cook approach, in particular by employing lattice topology in the inhibitory layer and using multi-level inhibition schemes. We demonstrate increased learning speed on machine learning tasks due to these architectural change. Next, we describe the PyTorch-based BindsNET \footnote{\texttt{BindsNET} code is available at \texttt{https://github.com/Hananel-Hazan/bindsnet}} simulator \cite{10.3389/fninf.2018.00089, hazanetal18}, where we were able to replicate previous results using the BRIAN library and implement new experiments based on the greater flexibility of BindsNET. The employed testbeds include images from MNIST dataset \cite{lecun-mnisthandwrittendigit-2010} and images obtained by the Atari Breakout game. Using optimally tuned parameters, our approach produces performance improvements over previous state-of-the-art unsupervised spiking neural networks. 

Our work shows how to use networks of self-organizing spiking neurons to (1) simultaneously cluster and classify, and (2) operate on image data. We also shows how the training done without the need to know the size or the distribution of the input. Thus, this work open the feasibility of implementing spiking neural networks for efficient machine learning applications to be able to train on the fly in unsupervised manner. 

In this paper we show examples of SNN learned image filters and map formation and report test data performance based on labels determined with spiking activity from the training phase.

\section{Methods}
\label{s:methods}

This section summarizes the applied spiking neural network structure and learning algorithms using leaky integrate-and-fire (LIF) neurons and spike-timing-dependent plasticity (STDP), as initially outlined in \cite{p._u._diehl_unsupervised_2015}, extended in \cite{10.3389/fninf.2018.00089, saundersetal18}, and supplemented in \cite{hazanetal18} with inter-neuron distance-dependent inhibition strength.

\subsection{LIF neurons and STDP learning}

It has been demonstrated that, in regard to the number of neural units needed, networks of spiking neurons constitute a more computationally powerful class of neural network models than that comprised of traditional ANNs \cite{maass_networks_1997}. Namely, single spiking neurons are able to compute biologically relevant functions that would require many more ANN hidden units to realize. The basic computational operations of an ANN do not incorporate time, and unlike the all-or-nothing action potential of the biological neuron, they communicate by sending precise floating-point numbers downstream to the next layer of processing units. The standard unit used in traditional neural networks is synchronous; i.e., all neurons within a layer participate in computation at each step during forward (\textit{inference}) and backward (\textit{learning}) passes, whereas spiking neurons are well-known for their asynchronous operation; i.e., they are updated only as needed. Moreover, neurons of ANNs do not have memories of their previous actions, compared to spiking neurons, which often integrate time naturally by keeping track of decaying memory traces or implementing leaky voltages (membrane potentials).

A computationally efficient choice of biologically-inspired unit of computation is the leaky integrate-and-fire (LIF) spiking neuron \cite{spiking}. These incorporate time in their operation by integrating a moderately quickly decaying voltage as simulation progresses, yet are simple enough to be incorporated in the large networks required for processing large amounts of images or other complex data. In our network architectures, we use these units; some are \textit{excitatory} (increasing the potentials of their downstream neighbors) and others are \textit{inhibitory} (decreasing the potentials of their downstream neighbors). Synaptic conductances are modeled, which are used in the update equation for neuron membrane potential. The potentials are given by

\begin{equation}
\tau \frac{dv}{dt} = (v_{rest} - v) + g_e (E_{exc} - v) + g_i (E_{inh} - v),
\end{equation}

where $v_{rest}$ is the resting membrane potential, $E_{exc}, E_{inh}$ are the equilibrium potentials of excitatory and inhibitory synapses, respectively, and $g_e, g_i$ are the conductances of excitatory and inhibitory synapses, respectively. When a neuron's membrane potential exceeds a threshold $v_{thresh}$, it fires an action potential and resets back to $v_{reset}$. The neuron then undergoes a short (5ms) refractory period, during which time it does not integrate input and cannot spike.

Synapses are modeled by conductance changes and synaptic weights $w$: a synapse increases its conductance at the moment a pre-synaptic action potential arrives by $w$ [6]. Otherwise, the conductances are decaying exponentially. The dynamics of the synaptic conductance are given by

\begin{equation}
\tau_{g_e} \frac{dg_e}{dt} = -g_e, \tau_{g_i} \frac{dg_i}{dt} = -g_i.
\end{equation}


Spike-timing-dependent plasticity \cite{Markram1997RegulationOS, bi_synaptic_1998} is used to modify the weights of synapses connecting certain neurons. We call the neuron from which a synapse projects \textit{pre-synaptic}, and the one to which it connects \textit{post-synaptic}. For the sake of computational efficiency, we use \textit{online STDP}, in which spike traces $x_{pre}$ and $x_{post}$ are recorded for every synapse, a decaying memory of its recent pre- and post-synaptic spiking history. Each time a pre-synaptic (post-synaptic) spike occurs, $x_{pre}$ ($x_{post}$) is set to 1; otherwise, it decays exponentially to zero with a time constant $\tau_{trace}$. When a pre- or post-synaptic spike occurs, a weight change $\Delta w$ is computed using a STDP update rule. The rule we chose for our experiments is given by

\begin{gather}
\Delta w = 
\begin{cases}
	\eta_{post} x_{pre} (w_{max} - w) & \text{on post-synaptic spike} \\
    - \eta_{pre} x_{post} w & \text{on pre-synaptic spike}
\end{cases}
\end{gather}

The term $\eta$ denotes a learning rate and $w_{max}$ is the maximal allowed synaptic weight.

\subsection{SNN architecture and operating principle}

The basic SNN of \cite{p._u._diehl_unsupervised_2015} has a multi-layer structure, comprised of an input layer, an excitatory layer, and an inhibitory layer. For image data, the input layer is an array of size $n_{\textrm{n\_input}} = h \times w$, where $h, w$ are the image's vertical and horizontal span in pixels, respectively. The excitatory and inhibitory layers have arbitrary size $n_\textrm{n\_neurons}$. The input layer is connected all-to-all with the excitatory neurons, the excitatory layer is connected one-to-one with the inhibitory layer, and each inhibitory neuron connects to all excitatory neurons, except the one from which it receives its one-to-one connection. This \textit{lateral inhibition} creates a competition between excitatory neurons, in the sense that when an excitatory neuron fires and activates its corresponding inhibitory neuron, all other excitatory neurons are inhibited, a mechanism akin to the operation of a \textit{winner-take-all} (WTA) circuit \cite{Oster:2009:CSW:1596917.1596919}. Spikes are generated in the input neurons by converting each pixel in an image into a Poisson spike train \cite{w._gerstner_spiking_2002} with firing rate $\lambda$ proportional to its intensity; i.e., more intense pixels are converted into spike trains with a higher average firing rate. This process is carried out for a specific simulation time, and the STDP rule is computed for the connections between the input and excitatory layers. It is an important property of the architecture that once a neuron fires, it inhibits all other neurons according to the inhibition scheme described above. For further details on the basic model structure, including considerations on system size, see \cite{p._u._diehl_unsupervised_2015, saundersetal18}.

\begin{figure}
  \centering
  \includegraphics[width=0.55\textwidth, height=6cm]{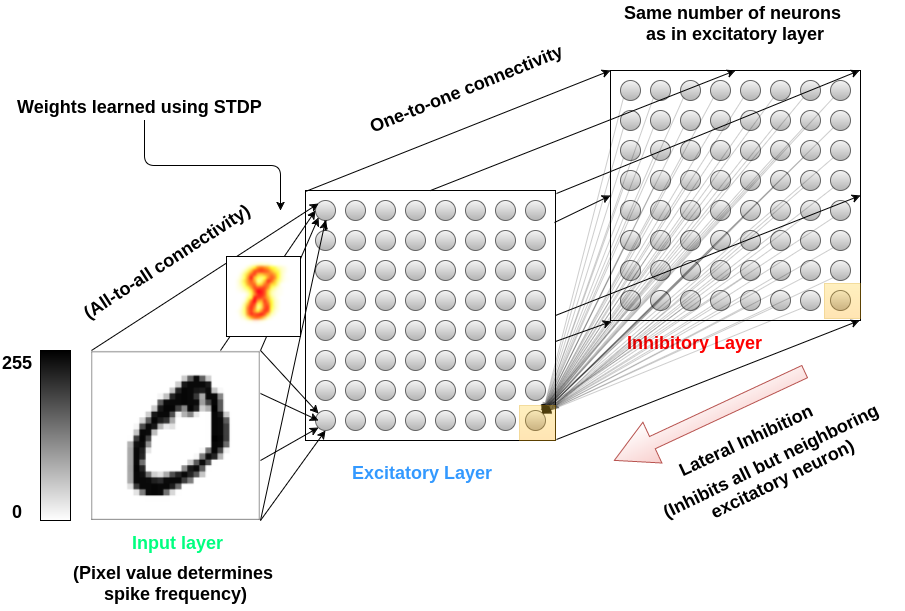}
  \includegraphics[width=0.4\textwidth, height=6cm]{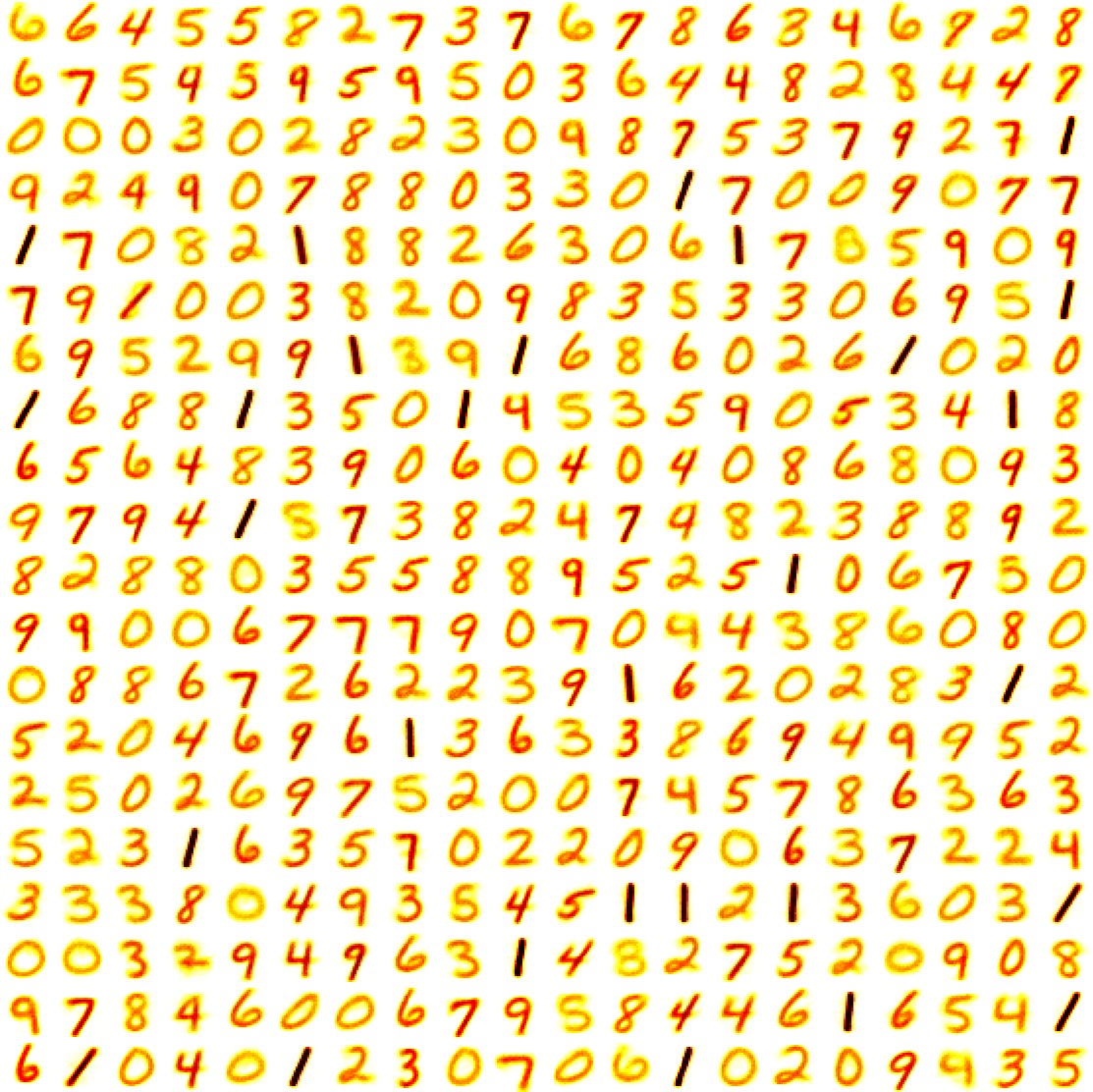}\\
  (a)\hspace{7cm}(b)
  \caption{Illustration of the SNN structure and its operation; (a) basic spiking neural network architecture, following 
  \cite{p._u._diehl_unsupervised_2015}; (b) learned filters from baseline model.}
  \label{fig:baseline_architecture}
\end{figure}

The WTA-like mechanism described above is what allows individual neurons to learn unique filters. Increasing the number of neurons in the excitatory and inhibitory layers has the effect of enabling a SNN to \textit{memorize} more canonical examples from the training data, and therefore recognize more patterns during the test phase. This network architecture is depicted in Figure \ref{fig:baseline_architecture}a, and an example set of filters learned from the MNIST handwritten digit dataset \cite{lecun-mnisthandwrittendigit-2010} is shown in Figure \ref{fig:baseline_architecture}b. Although the filters are arranged in a two-dimensional matrix, there is no neighborhood structure between the learned filters; i.e., neighboring excitatory neuron filters do not have any particular relationship with each other.

\begin{figure}[H]
  \centering
  \includegraphics[width=0.6\textwidth]{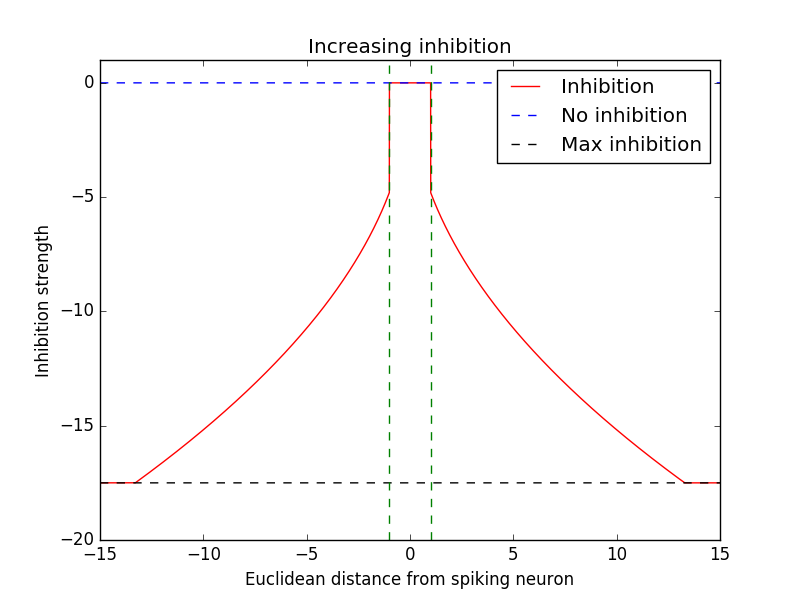}
  \caption{Inhibition as a function of Euclidean distance}
  \label{fig:inhibition_function}
\end{figure}

\subsection{Introducing topology in layers by increasing inhibition with inter-neuron distance}

Parameters for the spatial and temporal variation of inhibition are summarized in Table \ref{tab:temp_params}.

\begin{table}[]
  \centering
    \caption{Definition of parameters used in training of LM-SNNs.}
    \begin{tabular}{|l|l|}
        \toprule
        $c_\textrm{inhib}$ & Factor to scale the distance between neurons to determine inhibition strength \\
        $c_\textrm{min}$ & Minimum allowable inhibition strength \\
        $c_\textrm{max}$ & Maximum allowable inhibition strength \\
        $p_\textrm{low}$ & Proportion of training where $c_\textrm{inhib}$ is fixed to $c_\textrm{min}$ \\
        $p_\textrm{grow}$ & Proportion of training where $c_\textrm{inhib}$ is linearly increased from $c_\textrm{min}$ to $c_\textrm{max}$ \\
        \bottomrule
    \end{tabular}
    \label{tab:temp_params}
\end{table}

\subsubsection{Spatial profile of inhibition}

We introduce a change to the foregoing SNN architecture, by defining a profile in inter-neuron inhibition as a function of their distance on a lattice. Instead of inhibiting all other neurons at a large fixed rate as in \cite{saundersetal18, p._u._diehl_unsupervised_2015}, we increase the level of inhibition with the square root of the Euclidean distance between neurons, similar to the SOM learning algorithm:

\begin{gather}
    inhib_{i, j} = 
    \begin{cases}
        c_\textrm{inhib} \sqrt{(x_i - x_j)^2 + (y_i - y_j)^2}, & \textrm{ if } c_\textrm{inhib} \sqrt{(x_i - x_j)^2 + (y_i - y_j)^2} < c_\textrm{max}, \\
        c_\textrm{max}, & \textrm{otherwise},
    \end{cases}
\end{gather}

where $i, j$ indexes a pair of neurons, and $(x_i, y_i), (x_j, y_j)$ give their position on a two-dimensional lattice. This assumes that the neurons in the excitatory layer are arranged as such. It requires a parameter $c_\textrm{inhib}$, which is multiplied by the distance to compute inhibition levels, as well as a parameter $c_\textrm{max}$ that gives the maximum allowed inhibition; see Figure \ref{fig:inhibition_function}. With proper choice of $c_\textrm{inhib}$, when a neuron exceeds its firing threshold $v_\textrm{threshold}$, instead of inhibiting all other neurons from firing, a neighborhood of nearby neurons will be weakly inhibited and may have the chance to fire. This encourages neighborhoods of neurons to fire for the same inputs and learn similar filters. The radius of the neighborhood is determined by the $c_\textrm{inhib}$ parameter.

This inhibition profile is inspired by the self-organizing map (SOM) \cite{Kohonen1982} and it is included in part to reduce the degree of competition imposed by the connections from the inhibitory layer. The original inhibitory scheme introduced in \cite{p._u._diehl_unsupervised_2015} produces a network activation which is typically too sparse during training to learn filters quickly. Namely, after an initial transient following the introduction of an input example, one neuron fires first and causes all others to become strongly inhibited and therefore quiescent. Additionally, the introduced profile causes excitatory neuron filters to self-organize into distinct clusters by similarity, which often correspond to categories of the input data.

In contrast to the SOM learning algorithm, this method maintains a single learning rate throughout the training process and is able to continue to learn if more data become available. In the SOM, the learning rate is decreasing throughout the training phase, enabling the network to change dramatically in early stages of the training and then gradually stabilize and fine-tune its weights by the end of the training phase. For clarity, lattice map spiking neural networks with all variants of increasing inhibition strategies are denoted as LM-SNNs.

\subsubsection{Temporal variation of inhibition profile}

\begin{figure}[H]
  \centering
  \subfloat[]{
  \includegraphics[width=0.43\textwidth]{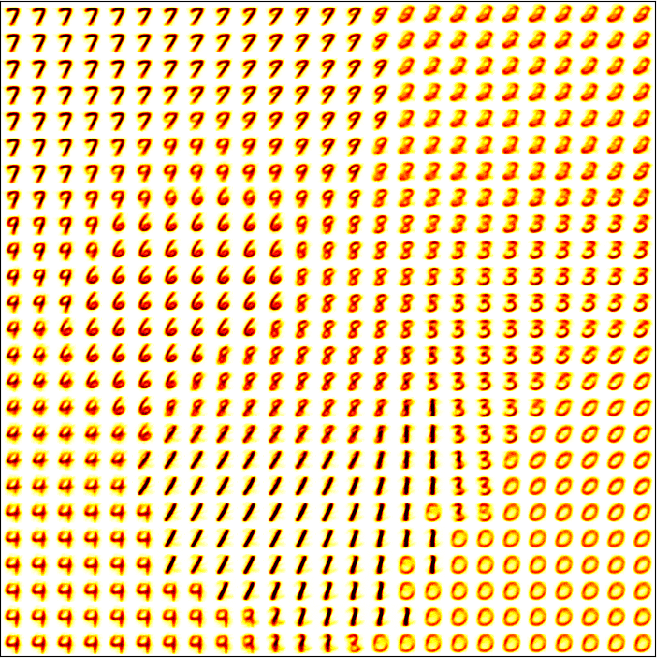}
  \label{fig:increasing_filters}}
  \subfloat[]{
  \includegraphics[width=0.48\textwidth]{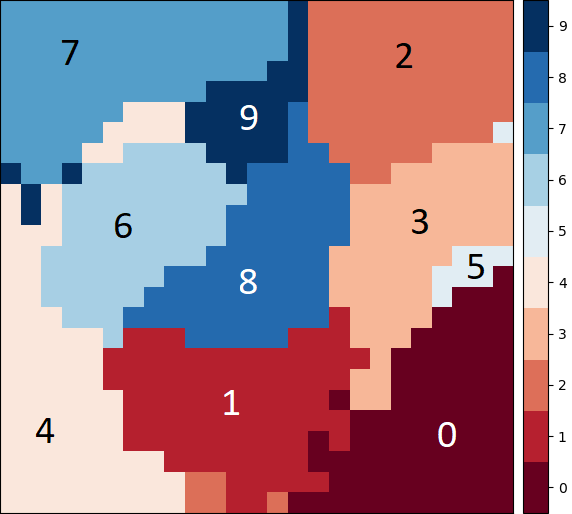}
  \label{fig:increasing_assignments}}
  \caption{Illustration of LM-SNN with increasing inhibition algorithm after training on the MNIST handwritten digit dataset; $c_{inhib}$ = 1.0; (a) 25x25 lattice of neuron filters; (b) class assignment labels are shown by color maps.}
  \label{fig:increasing_figure}
\end{figure}

\begin{figure}[H]
  \centering
  \subfloat[25x25 lattice of neuron filters]{
  \includegraphics[width=0.43\textwidth,height=6.5cm,keepaspectratio]{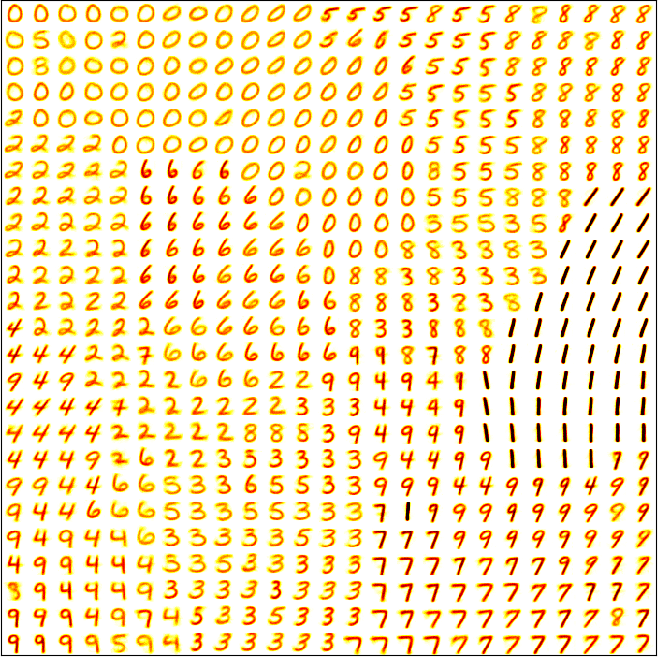}
  \label{fig:two_level_filters}}
  \subfloat[Neuron class assignments]{
  \includegraphics[width=0.48\textwidth,height=6.5cm,keepaspectratio]{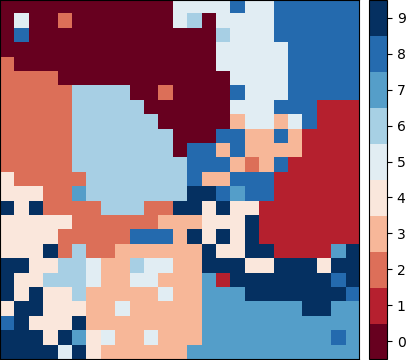}
  \label{fig:two_level_assignments}}
  \caption{Illustration of LM-SNN with two-level inhibition algorithm using the MNIST handwritten digit dataset; 
(a) 25x25 lattice of neuron filters; (b) class assignments are shown by color maps.}
  \label{fig:two_level}
\end{figure}

\begin{itemize}
\item
\textit{Growing inhibition over training}

~We want to produce \textit{individualized} filters as learned by the SNN presented in \cite{p._u._diehl_unsupervised_2015}, yet retain the clustering of filters achieved by our \textit{increasing inhibition} modification. Distinct filters are necessary to ensure that our learned representation contains as little redundancy as possible, making the best use of model capacity. To that end, we implemented another modification to the inhibition scheme, where the inhibition constant $c_{\textrm{inhib}}$ grows on a linear schedule from a small value $c_{\textrm{min}} \approx 0.1$ to a large value $c_{\textrm{max}} \approx 17.5$. The \textit{increasing inhibition} strategy is used as before; however, by the end of network training, the inhibition level is equivalent to that of \cite{p._u._diehl_unsupervised_2015}. 
To illustrate the effect of the spatial profile of inhibition, we show Figure \ref{fig:increasing_figure}, where increasing inhibition creates clusters of filters in the excitatory layers; the corresponding layers are displayed as well. Compare this to Figure \ref{fig:baseline_architecture}b, in which filters are learned by using fixed, uniform inhibition across the space, according to the learning algorithm described in \cite{p._u._diehl_unsupervised_2015}.

In Figure \ref{fig:increasing_figure}, the filters self-organize into smoothly-varying clusters, and then individualize as the inhibition level becomes large. We also consider growing the inhibition level to $c_{\textrm{max}}$ for some percentage of the training ($p_{\textrm{grow}}$) and holding it fixed for the rest ($1 - p_{\textrm{grow}}$). During the last $1 - p_{\textrm{grow}}$ percentage of the training phase, neuron filters are allowed to individualize more, enabling the network to represent less frequent examples from the training data.

\item
\textit{Two-level inhibition}
~To remove the need to re-compute inhibitory synapse strengths throughout network training, we implemented a simple \textit{two-level inhibition} scheme: For the first $p_{\textrm{low}}$ proportion of the training, the network is trained with inhibition level $c_{\textrm{min}}$; for the last $1 - p_{\textrm{low}}$ proportion, the network is trained with $c_{\textrm{max}}$. The inhibition level is not smoothly varied between the two levels, but jumps suddenly at the $p_{\textrm{low}}$ mark. At the low inhibition level, large neighborhoods of excitatory neurons compete \textit{together} to fire for certain types of inputs, eventually organizing their filters into a SOM-like representation of the dataset. At the high inhibition level, however, neurons typically maintain yet sharpen the filter acquired during the low inhibition portion of training. In this way, we obtain filter maps similar to those learned using the \textit{growing inhibition} mechanism, but with a simpler implementation. This inhibition strategy represents a middle ground between that of \cite{p._u._diehl_unsupervised_2015} and our \textit{increasing inhibition} scheme.

See Figure \ref{fig:two_level} for an example of learned filter map and neuron class assignments with two-level inhibition. There is some degree of clustering of the filters; however, as the inhibition level approaches that of \cite{p._u._diehl_unsupervised_2015}, they may eventually move away from the digit originally representing their weights, producing a somewhat fragmented geometry of the filter clustering. The degree of this fragmentation depends on the choice of $p_\textrm{low}$. Namely, with more time training with the high $c_\textrm{max}$ inhibition level, the more likely a neuron is to change its filter to represent data outside of its originally converged class.

\end{itemize}

\subsection{Reading out learned representations from the activations of SNN}

Due to the unsupervised nature of training in LM-SNN, the known class labels of input data are not utilized at that stage. The fact that the class labels are not used in the training process leads to a decreased classification performance, no matter what read-out scheme is used after the training. This may seem a waste of resources, however, it may become a virtue if the class labels are not known or unreliable. Accordingly, the application domain of SNNs is clearly different that of massive Deep Learning NNs trained by supervised gradient descent. Nevertheless, we may wish to evaluate the quality of the representations the SNN has learnt in terms of classification performance. 

There are several ways to achieve this goal, which is dependent on the readout scheme employed. An efficient readout scheme can be based on the observation that the network's representation of the dataset is encoded in the learned weights of synapses connecting the input and excitatory layers and the adaptive threshold parameters of the excitatory neurons. We use the activity of the neurons in the excitatory layer with their filter weights held fixed to accomplish these tasks.
We perform a two-step procedure based on individual neuron activations: (1)
 \textit{Labeling:} Excitatory neurons receive a label corresponding to the category they represent; 
 (2) \textit{Classifying:} New examples are classified based on the spiking activity of neurons with the previously attached labels.
 
There are many popular labeling and classification schemes, some of the listed below, which we will test.
The first 4 can be considered as rate-based schemes; namely the timing of individual neurons' spikes are discarded in favor of a rate-based code, see, e.g.,  \cite{Gollisch1108}. Some voting schemes consider the information contained in the timing, which is the case of the last option listed below:
\begin{enumerate}
\item
\textit{All voting:} Neurons are assigned the label of the input class for which they have fired most on average, and these labels are used to classify new data based on network activation. 
\item
 \textit{Confidence weighting:} In this voting scheme, we record the proportions of spikes each neuron fires for each input class, and used a weighted sum of these proportions and network activity to classify new inputs. 
 \item
 \textit{Distance-based:} New data is labeled with the class label of the neuron whose filter most closely matches the input data as measured by Euclidean distance. This is akin to the $k$-nearest neighbors algorithm with $k$ = 1, where the stored data examples are represented by the neurons' excitatory filter weights.
 \item
 \textit{n-gram:} In order to leverage information contained in the timing (ordering) of excitatory spikes, we used a $n$-gram based classification scheme, in which the timing and the order of spikes is used to make a prediction. This algorithm, inspired by rank order coding  \cite{thorpe1998rank, delorme2001networks}, represents an approach different from the scheme above. This method is able to utilize the pattern and the timing of first individual spikes in order to make predictions without waiting for a fixed amount of information. Therefore, this method can be trained on-line (over the course of training) and can be adapted incrementally.
 
\end{enumerate}

In Section \ref{sec:Results}, we evaluate networks with the \textit{all}, \textit{confidence weighting}, \textit{distance} and \textit{n-gram} voting schemes. 
Note that the $n-gram$ scheme, like the rate-based schemes, also follows a two-step procedure: a learning phase to estimate the $n-gram$ conditional class probabilities from the training data, and a classification phase where $n-grams$ are identified in the output spiking sequence and used to ``vote'' for the class they code for. 
$n-grams$ have long been used for sequential modeling, particularly in computational linguistics. There is evidence \cite{PETERSEN2001503} that most of the information in cortical neural networks is contained in the timing of individual spikes, and in particular on the importance of the exact timing of the first spike. Moreover, $n$-grams are able to identify repeating motifs \cite{RAICHMAN200896} in the activation of synchronized bursts in cultured neural networks and can also classify stimuli based on the time to first spike \cite{Kermany9588}.

\section{Results}
\label{sec:Results}

\subsection{Classification of MNIST images}

Here we summarize results obtained with the LM-SNN models using MNIST data;  for additional details, see \cite{saundersetal18, hazanetal18}. We omit results on the \textit{increasing} and \textit{growing inhibition} strategies in favor of results using the simpler \textit{two-level inhibition} strategy.
A large compilation of results demonstrates that network performance is robust to a wide range of hyper-parameter choices. This strategy is compared with a baseline SNN taken from \cite{p._u._diehl_unsupervised_2015},
and is shown to outperform it, especially in the regime of small networks, and particularly so with the \textit{n-gram} voting scheme. 
We discuss the robustness of LM-SNNs, showing a graceful degradation in performance with respect to random deletion of neurons and synapses.


\subsubsection{Comparison of Diehl $\&$ Cook baseline SNN and LM-SNN}

The networks comprising 625 excitatory and inhibitory neurons are trained on a single pass over the training data and evaluated on all 10K test examples. Test accuracy results and single standard deviations are reported in Table \ref{tab:two_level_accuracy}, each averaged over 5 independent trials. Results are demonstrated for a number of choices of parameters $p_{\textrm{low}}$, $c_{\textrm{min}}$, and $c_{\textrm{max}}$. The best performances are over $92\%$ for some hyperparameter configurations, which measures up to the Diehl $\&$ Cook original results. 

\begin{figure}[H]
  \centering
  \subfloat[]{
  \includegraphics[width=0.31\textwidth]{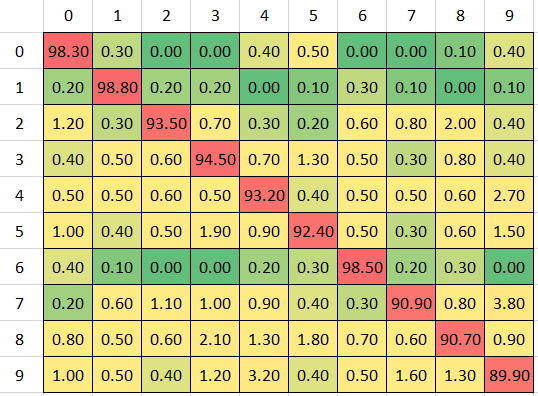}
  \label{fig:confusion_matrix_best}}
  \subfloat[]{
  \includegraphics[width=0.31\textwidth]{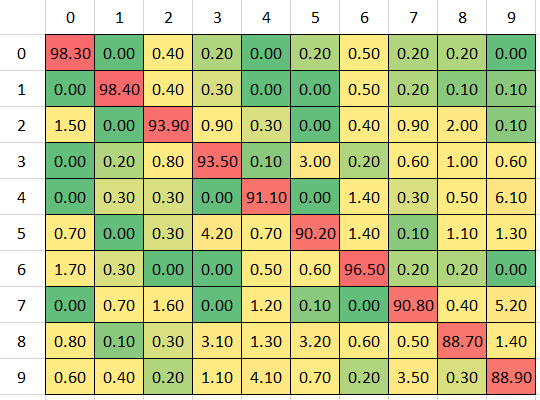}
  \label{fig:confusion_matrix_sec_best}}
  \subfloat[]{
  \includegraphics[width=0.31\textwidth]{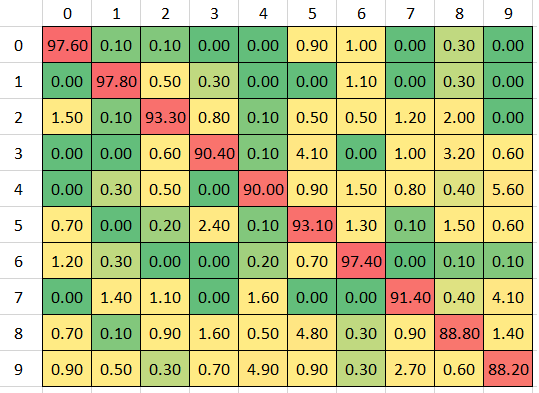}
  \label{fig:confusion_matrix_worst}}
  \caption{
  Average confusion matrices for networks with 1,600 neurons, based on the results in Table \ref{tab:eth_vs_two_level}, for the best classification scheme (Figure \ref{fig:confusion_matrix_best}; $n-gram$; 94.07$\%$), second best (Figure \ref{fig:confusion_matrix_sec_best}; distance; 93.03$\%$), and the worst (Figure \ref{fig:confusion_matrix_worst}; baseline Diehl $\&$ Cook; 92.80$\%$). Entries on row $i$ and column $j$ give the percentage of test examples that are classified as $j$, despite actually belonging to class $i$.
  }
  \label{fig:confusion_matrix}
\end{figure}

For a detailed comparison of our approach and the Diehl $\&$ Cook baseline results \cite{p._u._diehl_unsupervised_2015}, we conducted a systematic study of network sizes and parameters. Table \ref{tab:eth_vs_two_level} introduces a comparison of Diehl $\&$ Cook accuracies (baseline SNN in column 2) over several settings of the number of neurons in LM-SNN, using the \textit{confidence}, \textit{all}, \textit{distance} and \textit{n-gram} voting schemes. The two-level inhibition hyper-parameters are fixed to $p_{\textrm{low}} = 0.1, c_{\textrm{min}} = 1.0, c_{\textrm{max}} = 20.0$. For the \textit{n-gram} scheme, 12K examples are used for the learning phase. 
The smaller networks were trained for a single pass, while we conducted a 3-pass training for the 1,225 and 1,600 networks to explore peak accuracies. Detailed results with all single passes are given in \cite{hazanetal18}. To further evaluate the classification ability that is proposed in this paper, we compute average confusion matrices (Figure \ref{fig:confusion_matrix}) for networks with 1,600 neurons from Table \ref{tab:eth_vs_two_level}, for the best classification scheme ($n-gram$; 94.07$\%$), the second-best (distance; 93.03$\%$), and the worst (baseline Diehl \& Cook; 92.80$\%$).

\begin{table}
\begin{center}
\caption{Test accuracy on MNIST images \\(two-level inhibition; $n_e, n_i$ = 625)}
\label{tab:two_level_accuracy}
\resizebox{\textwidth}{!}{
\begin{tabular}{||c|c|c|c|c|c||c|c|c|c|c|c||}
\toprule
$p_{\textrm{low}}$ & $c_{\textrm{min}}$ & $c_{\textrm{max}}$ & \textit{distance} &      \textit{all} & \textit{confidence} & $p_{\textrm{low}}$ & $c_{\textrm{min}}$ & $c_{\textrm{max}}$ & \textit{distance} &      \textit{all} & \textit{confidence} \\
\midrule
0.1 &                0.1 &               15.0 &   91.8 $\pm$ 0.63 &   91.4 $\pm$ 0.13 &    91.69 $\pm$ 0.63 &			  0.25 &                0.1 &               15.0 &  91.51 $\pm$ 0.25 &  90.62 $\pm$ 0.26 &    90.97 $\pm$ 0.25 \\
0.1 &                0.1 &               17.5 &  92.02 $\pm$ 0.38 &  91.26 $\pm$ 0.11 &    91.68 $\pm$ 0.38 &              0.25 &                0.1 &               17.5 &  91.83 $\pm$ 0.18 &  91.06 $\pm$ 0.18 &    91.54 $\pm$ 0.18 \\
0.1 &                0.1 &               20.0 &  92.38 $\pm$ 0.49 &  91.54 $\pm$ 0.14 &     92.1 $\pm$ 0.49 &              0.25 &                0.1 &               20.0 &  92.16 $\pm$ 0.34 &  91.07 $\pm$ 0.29 &    91.83 $\pm$ 0.34 \\
0.1 &                1.0 &               15.0 &  91.67 $\pm$ 0.63 &  91.12 $\pm$ 0.39 &    91.59 $\pm$ 0.63 &              0.25 &                1.0 &               15.0 &  91.36 $\pm$ 0.24 &  90.26 $\pm$ 0.21 &    90.77 $\pm$ 0.24 \\
0.1 &                1.0 &               17.5 &  92.25 $\pm$ 0.42 &  91.32 $\pm$ 0.29 &    91.67 $\pm$ 0.42 &              0.25 &                1.0 &               17.5 &  91.78 $\pm$ 0.61 &  91.09 $\pm$ 0.29 &    91.46 $\pm$ 0.61 \\
0.1 &                1.0 &               20.0 &  92.36 $\pm$ 0.66 &  91.44 $\pm$ 0.38 &     91.9 $\pm$ 0.66 &              0.25 &                1.0 &               20.0 &   \textbf{92.3 $\pm$ 0.19} &  91.51 $\pm$ 0.17 &    91.98 $\pm$ 0.19 \\
0.1 &                2.5 &               15.0 &  91.99 $\pm$ 0.53 &  91.01 $\pm$ 0.29 &     91.3 $\pm$ 0.53 &              0.25 &                2.5 &               15.0 &  91.65 $\pm$ 0.64 &  90.89 $\pm$ 0.31 &    91.19 $\pm$ 0.64 \\
0.1 &                2.5 &               17.5 &  92.17 $\pm$ 0.39 &  91.49 $\pm$ 0.17 &    91.86 $\pm$ 0.39 &              0.25 &                2.5 &               17.5 &  92.04 $\pm$ 0.62 &  91.52 $\pm$ 0.27 &    91.95 $\pm$ 0.62 \\
0.1 &                2.5 &               20.0 &  \textbf{92.55 $\pm$ 0.54} &   92.07 $\pm$ 0.3 &    92.49 $\pm$ 0.54 &              0.25 &                2.5 &               20.0 &  92.26 $\pm$ 0.33 &  91.43 $\pm$ 0.48 &    91.97 $\pm$ 0.33 \\
\bottomrule
\end{tabular}}
\end{center}
\end{table}

Tests showed that bi-grams gave the best performance, and hence we fixed $n = 2$. 5 independent experiments with different initial configurations and Poisson spike trains were run, and their results are averaged and reported along with a single standard deviation. 
While the \textit{confidence}, \textit{all} and \textit{n-gram} schemes use the activity of the network in order to classify new data, the \textit{distance} scheme simply labels new inputs with the label of the neuron whose filter most closely matches the input. This last evaluation scheme is reminiscent of the one-nearest neighbor algorithm. However, our spiking neural networks learn prototypical data vectors, whereas the one-nearest neighbor method stores the entire dataset to use during evaluation.
The results for networks with 1,225 and 1,600 neurons suggest either or both of (1) the training algorithm for both of the baseline SNN and the LM-SNN does not make appropriate use of all data on a single training pass, or (2) network capacity is too large to adequately learn all filter weights with a single pass over the data. Inspection of the convergence of filter weights during training (data not shown) suggests that the training algorithm needs to be adjusted for greater data efficiency. The fact that training with more epochs improves accuracy also points to the fact that network capacity may be too high to fit with a single pass through the data.

\begin{table}
    \begin{center}
    \caption{Baseline SNN (Diehl $\&$ Cook) vs. LM-SNN \\(two-level inhibition; variable $n_e, n_i$ )}
    \label{tab:eth_vs_two_level}
    \resizebox{\textwidth}{!}{
    \begin{tabular}{||c|c|c|c|c|c|c||}
        \hline
        $n_e$, $n_i$ & Baseline D $\&$ C & LM-SNN (\textit{confidence}) & LM-SNN (\textit{all}) & LM-SNN (\textit{distance}) & LM-SNN (\textit{n-gram}) \\ [0.5ex]
        \hline \\ [-1.0em]
        100 & 80.71\% $\pm$ 1.66\% & 82.94\% $\pm$ 1.47\% & 81.12\% $\pm$ 1.96\% & 85.11\% $\pm$ 0.74\% & 85.71\% $\pm$ 0.85\% \\
        225 & 85.25\% $\pm$ 1.48\% & 88.49\% $\pm$ 0.48\% & 87.33\% $\pm$ 0.59\% & 89.11\% $\pm$ 0.37\% & 90.50\% $\pm$ 0.43\% \\
        400 & 88.74\% $\pm$ 0.38\% & 91\% $\pm$ 0.56\% & 90.56\% $\pm$ 0.67\% & 91.4\% $\pm$ 0.38\% & 92.56\% $\pm$ 0.48\% \\ 
        625 & 91.27\% $\pm$ 0.29\% & 92.14\% $\pm$ 0.50\% & 91.69\% $\pm$ 0.59\% & 92.37\% $\pm$ 0.29\% & 93.39\% $\pm$ 0.25\%\\
        900 & 92.63\% $\pm$ 0.28\% & 92.36\% $\pm$ 0.63\% & 91.73\% $\pm$ 0.7\% & 92.77\% $\pm$ 0.26\% & 93.25\% $\pm$ 0.57\% \\
        1,225$^{a}$ & 92.43\% $\pm$ 0.23\% & 92.57\% $\pm$ 0.57\% & 91.85\% $\pm$ 0.52\% & 92.48\% $\pm$ 0.29\% & 93.87\% $\pm$ 0.25\% \\
        1,600$^{a}$ & 92.80\% $\pm$ 0.49\% & 92.96\% $\pm$ 0.56\% & 92.87\% $\pm$ 0.49\% & 93.03\% $\pm$ 0.30\% & \textbf{94.07\% $\pm$ 0.46\%} \\
        \hline
            \end{tabular}}
            \footnotesize{$^{a}$ Results for 1,225 and 1,600 nodes have been obtained using 3 passes through the MNIST training data.}
    \end{center}
\end{table}

\begin{figure}
  \centering
  \includegraphics[width=0.65\textwidth]{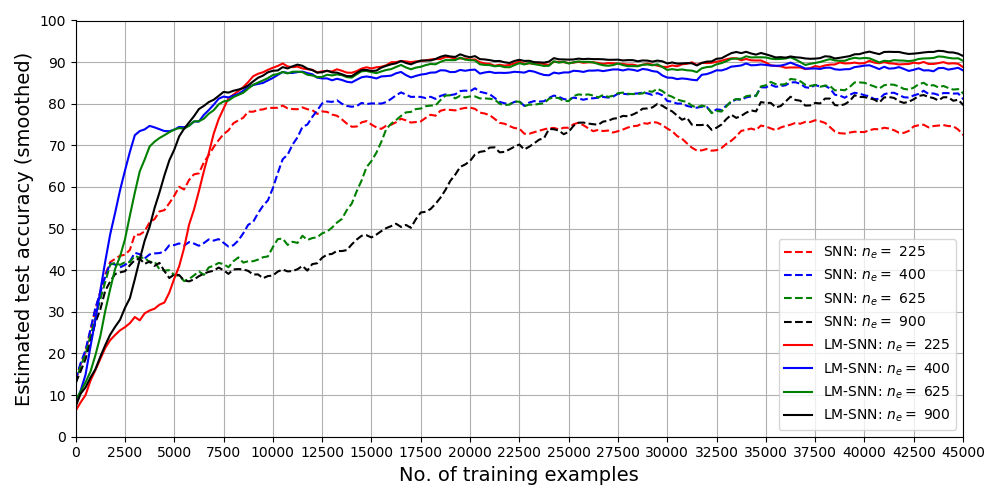}
  \caption{LM-SNN vs. Diehl $\&$ Cook SNN smoothed performance estimate over the training phase for MNIST data.}
  \label{fig:convergence}
\end{figure}
         

%
%

\subsubsection{LM-SNN convergence and online learning option}
\label{ssec:convergence}

A major advantage of using a relaxed inhibition scheme is the ability to learn a better data representation with fewer training examples. While training LM-SNNs using the \textit{growing} and \textit{two-level} inhibition strategies, we observe a convergence to optimal network performance well before SNNs trained with large, constant inhibition as in \cite{p._u._diehl_unsupervised_2015}. With small inhibition (increasing with the distance between pairs of neurons) the development of filters occurs quickly, allowing the fast attainment of decent performance. Over the course of the training, those filters are gradually refined as the inter-neuron inhibition grows and allows for the individualized firing of neurons.

We show in Figure \ref{fig:convergence} a comparison of the convergence in estimated test performance for networks of various sizes. D $\&$ C baseline SNNs are trained with large, constant inhibition $c_{inhib} = 20.0$, while LM-SNNs are trained with the \textit{two-level inhibition} strategy with parameter values $p_{\textrm{low}} = 0.1$, $c_{\textrm{min}} = 1.0$, and $c_{\textrm{max}} = 20.0$. Estimates are calculated by assigning labels to neurons based on the firing activity on 250 training examples, and using these labels along with the aforementioned classification methods to classify the next 250 training examples. Training accuracy curves are \textit{smoothed} by averaging each estimate with a window of the 10 neighboring estimates. The performance of LM-SNNs quickly outpace that of baseline SNNs, and obtain near-optimal accuracy after seeing between 10 and 20 thousand training examples, regardless of the number of excitatory and inhibitory neurons, in the range of 225 to 900 neurons. As the result of the employed spatially distributed inhibition strategy, larger LM-SNNs achieve better accuracy more quickly, due to their ability to learn more filters at once. On the other hand, SNNs are limited to learning one filter at a time, and their accuracy during the training phase is hindered by the size of the network as a result. The observed rapid learning property of LM-SNN points to the potential of their use under various online learning conditions. 

\subsubsection{Robustness of LM-SNN performance with sparse data}
\label{ssec:sparse}

To test the performance of the SNNs, we explored the impact of diminishing connection strengths on the learning process. Accordingly, instead of connecting the input data pixels fully with all the neurons in the excitatory layer, we experimented with varying degrees of random sparse connectivity. The applied procedure has been as follows. Before training, some percentage of synapses are randomly eliminated, after which the training and test phases proceeded as usual. We studied whether varying levels of sparsity make LM-SNN more robust to outliers in the MNIST data, therefore increasing the chance of good test performance. We study how does the network perform in the event of missing features, whether it exhibits graceful degradation in performance as the input data becomes less clear.

Table \ref{tab:sparse_results} displays accuracy results with varying level of sparsity of connections to the input as defined above, based on averaging ten independent training and test runs, reported along with their standard deviations. Interestingly, small amounts of sparsity do not significantly degrade the network performance. Moreover, even with nearly all connections removed, the network maintains reasonable accuracy. In particular, removing 90$\%$ of synapses from the input, networks of 625 excitatory and inhibitory neurons achieve nearly 60\% test accuracy after being trained using the \textit{two-level inhibition} mechanism. This result demonstrates the robustness of our system to missing information in the input data.

\begin{table}[H]
\begin{center}
 \caption{Sparse input test accuracy}
 \label{tab:sparse_results}
 \begin{tabular}{||c|c|c||}
 \hline
 $n_e$, $n_i$ & \% sparsity & Test accuracy \\ [0.5ex] 
 \hline
 625 & 0\% & \textbf{91.71}\% $\pm$ 0.23\% \\
 625 & 10\% & 91.48\% $\pm$ 0.31\% \\
 625 & 25\% & 89.79\% $\pm$ 0.66\% \\
 625 & 50\% & 85.83\% $\pm$ 0.95\% \\ 
 625 & 75\% & 75.71\% $\pm$ 1.20\% \\ 
 625 & 90\% & 58.60\% $\pm$ 1.31\% \\ 
 \hline
\end{tabular}
\end{center}
\end{table}

\subsection{Atari breakout frame classification}
\label{ssec:breakout}

\subsubsection{Description of Atari Breakout task domain}

\begin{figure}[H]
    \centering
    \includegraphics[width=0.30\textwidth,height=5cm]{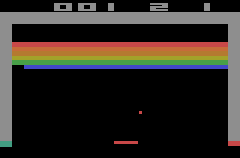}
    \caption{A frame from the Atari Breakout game. Note the pedal at the bottom, the colored bricks, and the ball (red) in the middle.}
    \label{fig:breakout}
\end{figure}

To demonstrate the flexibility of the LM-SNN model across different choices of task domains, we assembled a dataset of modified frames from the Atari game Breakout. 
We use the OpenAI Gym \cite{AIGYM16} to simulate the game environment. The game involves an image of 80x80 pixel on computer screen, which has a paddle at the bottom, a wall of multiple layers of bricks on the upper section, and a ball that moves over the screen. Atari Breakout frames are cropped to remove borders that appear to be irrelevant to the game-playing agent. An example game from the actual game in shown in Figure \ref{fig:breakout}. The game is played by moving a rectangular paddle horizontally across the bottom of the screen in order to keep a ball from entering the lowest part of the screen and being lost.  The possible actions in the Breakout game are "no action", "fire" (launch the ball from the paddle to start the round), "move right", and "move left". The game ends when all tiles from the top of the screen are cleared by the action of the bouncing ball. 

In this dataset, frames are labeled according to the corresponding action taken by a two-layer fully-connected artificial neural network (ANN) trained by the Q-learning algorithm; that is, a Deep Q-Network (DQN) \cite{mnih15}. The frames were created by taking a weighted sum of four consecutive in-game frames, in which the most recent frame has the highest weight, and the least recent the lowest weight.  Since some actions are taken more often than others, there is a class imbalance in the dataset we have curated. To mitigate this problem, we re-sample from each class-specific subset of data as many times as needed to build a new dataset of arbitrary size with balanced classes. In the original data, there are 6,850 examples, with approximately 49.3\% labeled "no action", 6.3\% labeled "fire", 28.4\% labeled "move right", and 16.0\% labeled "move left". We found through experiments that a re-sampling of 16,000 training examples (with 4,000 examples per class) saturated the learning ability of trained networks; i.e., training with more examples did not significantly improve classification accuracy.

We first modified the SNN architecture of \cite{p._u._diehl_unsupervised_2015} to use for classification of the Atari Breakout game frames. The input layer (after cropping out irrelevant regions) is 50 pixels high and 72 pixels wide, requiring a population of input neurons of size 50 $\times$ 72 = 3,600. The input population is connected \textit{all-to-all} to an output layer of neurons of arbitrary size $n_\textrm{neurons}$. We removed the ``inhibitory'' layer of neurons from the architecture of \cite{p._u._diehl_unsupervised_2015}, replacing it with a mechanistically similar inhibitory recurrent connection in the output layer; i.e., every output neuron connects to all other output neurons with inhibitory synapses. This allows us to simulate only $n_\textrm{input} + n_\textrm{neurons}$ instead of $n_\textrm{input} + 2*n_{neurons}$, reducing simulation time significantly while retaining the competitive inhibitory mechanism that enables both the baseline algorithm of \cite{p._u._diehl_unsupervised_2015} and our own method to learn.

Additionally, we re-implemented both the baseline SNN as well the LM-SNN network architectures in the BindsNET spiking neural networks simulation library \cite{10.3389/fninf.2018.00089}. Although the choice of software implementation is a minor detail, we found that this library enabled us to prototype and train networks more quickly than in the previously used simulation software \cite{goodman_brian_2009}.

To ensure that our changes didn't result in a reduction in network performance, we replicated the test accuracy results on the MNIST handwritten digits reported above (data not shown).

\subsubsection{Atari Breakout: Baseline results}
\label{ssec:baseline_breakout_results}

\begin{figure}[H]
    \centering
    \includegraphics[width=0.4\textwidth]{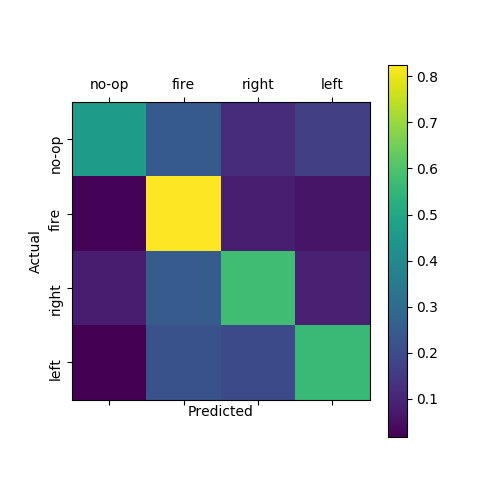}
    \caption{Test dataset confusion matrix for Atari Breakout frame classification using the network model of \cite{p._u._diehl_unsupervised_2015}. The network that produced this confusion matrix result achieved $\approx$60\% classification accuracy, with $\approx$50\% correct answers for ``no action'', $\approx$85\% for ``fire'', and $\approx$60\% for both ``move right'' and ``move left''. Axis labels are shortened for clarity of presentation.}
    \label{fig:breakout_confusion}
\end{figure}

\begin{figure}[H]
    \centering
    \includegraphics[width=0.45\textwidth]{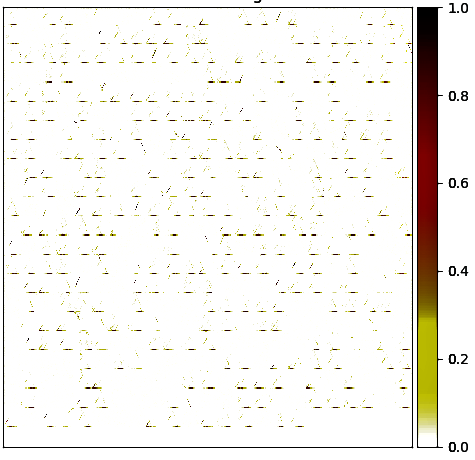}
    \includegraphics[width=0.51\textwidth]{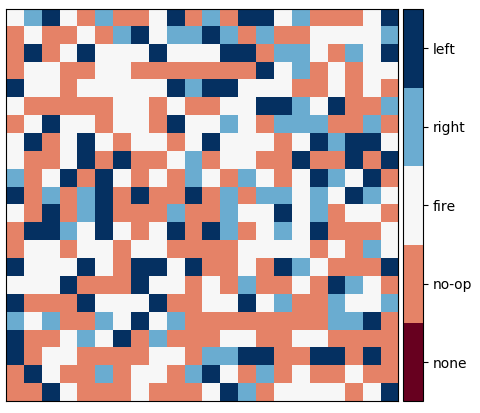}
    \caption{Atari Breakout network filter weights and corresponding class assignments. The contrast in the plot of filter weights is increased in order to show the learned weights a bit better. The filters clearly capture positions of the paddle (at the bottom of the frame) and the ball (which moves around the top portion of the frame), jointly or separately. Notice that some filters are not learned; i.e., some weights did not settle on a particular configuration of the game state.}
    \label{fig:breakout_weights_assignments}
\end{figure}

We applied the modified networks of \cite{p._u._diehl_unsupervised_2015} to the custom Atari game frame classification dataset. A typical confusion matrix produced by these networks is shown in Figure \ref{fig:breakout_confusion}, with corresponding weights and neuron class assignments in Figure \ref{fig:breakout_weights_assignments}.

Test classification results are shown in table \ref{tab:breakout_rebalance} for a number of configurations of hyper-parameters. All hyper-parameters are held fixed at sensible values except for $\theta^+$ (the amount to increase the neurons' firing thresholds post-spike) and $c_\textrm{norm}$ (the normalization constant for weights on the connection from input to output populations). For each configuration of hyper-parameters, the same experiment is run 5 times with different random seeds, and the average result with a single standard deviation is reported. As expected, the test classification accuracy typically increases with $n_\textrm{neurons}$, and we found that $\theta^+ = 5.0$ and $c_\textrm{norm}$ = 62.5 typically gave the best results. Varying other hyper-parameters and expanding the range of the search may lead to improved performance.


\begin{table}[H]
    \centering
    \footnotesize
    \begin{tabular}{||c|c|c|c|c|c||}
\toprule
$\theta^+$ & $c_\textrm{norm}$ & $n_\textrm{neurons}$ & \textit{all activity} & \textit{proportion weighting} & \textit{n-gram} \\
\midrule
1   & 50   & 100 & 53.69 $\pm$ 1.38 & 54.35 $\pm$ 0.73 & 54.70 $\pm$ 0.73 \\
    &      & 200 & 56.92 $\pm$ 0.30 & 57.15 $\pm$ 0.43 & 58.46 $\pm$ 0.43 \\
    &      & 300 & 57.25 $\pm$ 1.57 & 57.50 $\pm$ 1.73 & 57.65 $\pm$ 1.73 \\
    &      & 400 & 56.15 $\pm$ 1.84 & 56.13 $\pm$ 1.86 & 57.99 $\pm$ 1.86 \\
    &      & 500 & 56.80 $\pm$ 1.18 & 56.91 $\pm$ 1.10 & 58.03 $\pm$ 1.10 \\
    & 62.5 & 100 & 52.37 $\pm$ 1.53 & 52.58 $\pm$ 1.46 & 55.54 $\pm$ 1.46 \\
    &      & 200 & 56.05 $\pm$ 1.41 & 56.21 $\pm$ 1.48 & 57.21 $\pm$ 1.48 \\
    &      & 300 & 55.78 $\pm$ 1.66 & 56.07 $\pm$ 1.32 & 57.03 $\pm$ 1.32 \\
    &      & 400 & 56.47 $\pm$ 1.91 & 57.01 $\pm$ 2.10 & 57.67 $\pm$ 2.10 \\
    &      & 500 & 56.72 $\pm$ 2.04 & 57.20 $\pm$ 2.14 & 58.61 $\pm$ 2.14 \\
    & 75   & 100 & 53.84 $\pm$ 1.67 & 53.31 $\pm$ 2.26 & 55.42 $\pm$ 2.26 \\
    &      & 200 & 54.72 $\pm$ 2.54 & 54.71 $\pm$ 2.78 & 57.18 $\pm$ 2.70 \\
    &      & 300 & 56.76 $\pm$ 1.89 & 57.17 $\pm$ 1.92 & 57.75 $\pm$ 1.92 \\
    &      & 400 & 56.39 $\pm$ 1.55 & 56.66 $\pm$ 1.70 & 56.92 $\pm$ 1.70 \\
    &      & 500 & 57.18 $\pm$ 0.79 & 57.42 $\pm$ 0.74 & 58.08 $\pm$ 0.74 \\
\hline
2.5 & 50   & 100 & 53.62 $\pm$ 3.25 & 53.90 $\pm$ 3.24 & 55.33 $\pm$ 3.24 \\
    &      & 200 & 56.16 $\pm$ 2.03 & 56.47 $\pm$ 2.05 & 58.00 $\pm$ 2.05 \\
    &      & 300 & 58.07 $\pm$ 1.07 & 58.55 $\pm$ 0.94 & 59.38 $\pm$ 0.94 \\
    &      & 400 & 58.13 $\pm$ 1.23 & 58.30 $\pm$ 1.13 & 59.28 $\pm$ 1.13 \\
    &      & 500 & 57.05 $\pm$ 1.06 & 57.31 $\pm$ 1.32 & 59.14 $\pm$ 1.32 \\
    & 62.5 & 100 & 52.37 $\pm$ 4.54 & 52.84 $\pm$ 4.31 & 54.40 $\pm$ 4.31 \\
    &      & 200 & 57.79 $\pm$ 1.52 & 57.90 $\pm$ 1.82 & 59.00 $\pm$ 1.82 \\
    &      & 300 & 57.97 $\pm$ 1.20 & 58.35 $\pm$ 1.11 & 59.05 $\pm$ 1.11 \\
    &      & 400 & 57.64 $\pm$ 0.77 & 58.27 $\pm$ 0.43 & 59.15 $\pm$ 0.43 \\
    &      & 500 & 58.53 $\pm$ 0.40 & 58.86 $\pm$ 0.68 & 59.17 $\pm$ 0.68 \\
    & 75   & 100 & 52.34 $\pm$ 3.70 & 52.97 $\pm$ 3.70 & 53.31 $\pm$ 3.70 \\
    &      & 200 & 54.89 $\pm$ 4.85 & 55.21 $\pm$ 4.73 & 57.65 $\pm$ 4.73 \\
    &      & 300 & 58.34 $\pm$ 1.08 & 58.79 $\pm$ 1.19 & 60.14 $\pm$ 1.19 \\
    &      & 400 & 57.77 $\pm$ 1.54 & 58.29 $\pm$ 1.58 & 58.03 $\pm$ 1.58 \\
    &      & 500 & 57.85 $\pm$ 0.85 & 58.37 $\pm$ 1.20 & 58.74 $\pm$ 1.20 \\
\hline
5   & 50   & 100 & 54.58 $\pm$ 1.57 & 54.80 $\pm$ 2.18 & 55.06 $\pm$ 2.18 \\
    &      & 200 & 57.03 $\pm$ 3.61 & 57.44 $\pm$ 3.75 & 58.31 $\pm$ 3.75 \\
    &      & 300 & 58.31 $\pm$ 0.91 & 58.85 $\pm$ 0.88 & 59.39 $\pm$ 0.88 \\
    &      & 400 & 59.56 $\pm$ 1.02 & 60.11 $\pm$ 0.96 & 59.96 $\pm$ 0.96 \\
    &      & 500 & 59.89 $\pm$ 0.70 & 60.14 $\pm$ 0.80 & 59.68 $\pm$ 0.80 \\
    & 62.5 & 100 & 55.53 $\pm$ 1.77 & 55.78 $\pm$ 1.88 & 54.84 $\pm$ 1.88 \\
    &      & 200 & 56.90 $\pm$ 1.78 & 57.27 $\pm$ 1.85 & 58.05 $\pm$ 1.85 \\
    &      & 300 & 60.03 $\pm$ 1.24 & 60.48 $\pm$ 1.09 & 60.24 $\pm$ 1.09 \\
    &      & 400 & 58.68 $\pm$ 0.71 & 59.83 $\pm$ 0.54 & 60.09 $\pm$ 0.54 \\
    &      & 500 & 59.66 $\pm$ 0.96 & \textbf{60.59 $\pm$ 0.76} & 60.33 $\pm$ 0.76 \\
    & 75   & 100 & 55.61 $\pm$ 0.69 & 55.66 $\pm$ 0.55 & 54.60 $\pm$ 0.55 \\
    &      & 200 & 57.90 $\pm$ 1.39 & 58.30 $\pm$ 1.44 & 57.69 $\pm$ 1.44 \\
    &      & 300 & 59.28 $\pm$ 1.25 & 59.87 $\pm$ 1.17 & 59.37 $\pm$ 1.17 \\
    &      & 400 & 59.16 $\pm$ 1.38 & 59.46 $\pm$ 1.47 & 59.82 $\pm$ 1.47 \\
    &      & 500 & 59.69 $\pm$ 0.67 & 60.10 $\pm$ 0.24 & 60.15 $\pm$ 0.24 \\
    \bottomrule
    \end{tabular}
    \caption{Test performance results on the Atari breakout games frames classification task. Besides the reported variations of hyper-parameters, all other parameters are kept constant. 5 separate experiments are run, and averaged test results are reported along with standard deviations.}
    \label{tab:breakout_rebalance}
\end{table}

\subsubsection{Atari Breakout: LM-SNN results}
\label{ssec:lm-snn_breakout_results}

We used the lattice map spiking neural network (LM-SNN) architecture to jointly learn a clustering of filters and classify the Atari Breakout paired frame-actions dataset. We fix $\theta^+ = 5.0$ and $c_\textrm{norm} = 62.5$, and vary $n_{\textrm{neurons}}, c_{\textrm{start}}$ (starting inhibition level), and $p_\textrm{low}$ (proportion of training during which inhibition = $c_{\textrm{start}}$). As usual, after $n_{\textrm{low}} = p_\textrm{low} * n_\textrm{examples}$ training examples, the inhibition is switched to that of the baseline SNN \cite{p._u._diehl_unsupervised_2015}, and remains at this level for the rest of the training and test phases.

\begin{figure}[H]
    \centering
    \includegraphics[width=0.42\textwidth]{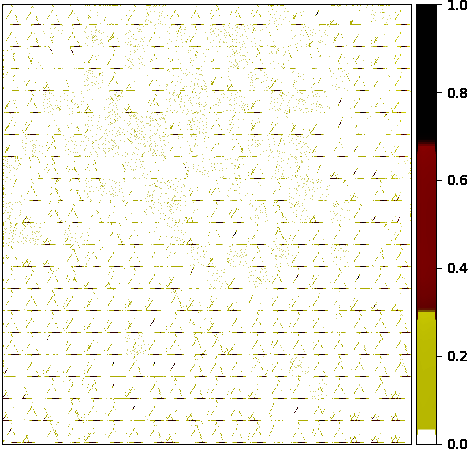}
    \includegraphics[width=0.48\textwidth]{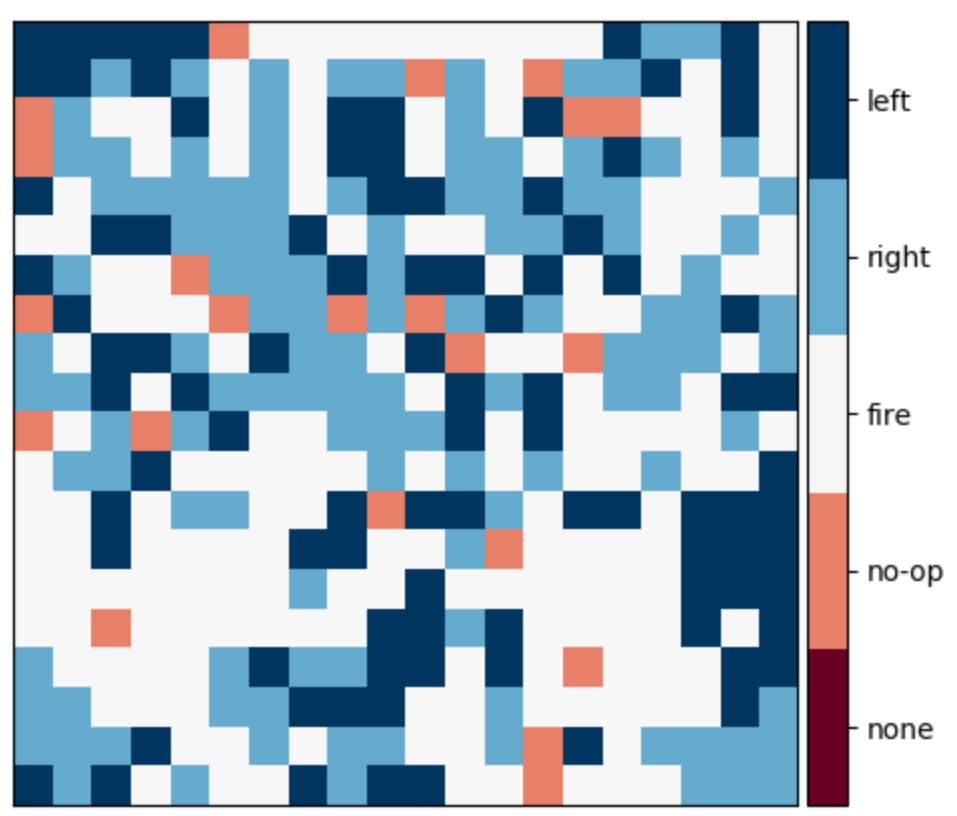}
    \caption{Example learned filters and corresponding class assignments from a two-level inhibition SNN with 400 output neurons and $p_\textrm{low} = 1.0$. Some filters remain un-settled, but many have learned a configuration of the ball and paddle. Though difficult to see, some filters appear to have clustered together on the basis of the similarity of their ball and paddle configurations. The visualized assignments reveal a degree of class label clustering, as well as the unbalanced learned representation of the training data.}
    \label{fig:breakout_two_level_filters}
\end{figure}

Examples of learned filters and corresponding label assignments from a network with $p_\textrm{low} = 1.0$ are plotted (with contrast levels adjusted for ease of viewing) in Figure \ref{fig:breakout_two_level_filters}. Some filters remain relatively un-settled, in that they remain close to their initialization weights (chosen uniformly at random in [0, 1], then normalized to sum to $c_\textrm{norm}$), while others learn a in-game configuration corresponding to certain frames from the training data. Some clustering of ball and paddle configurations is evident, but due to the nature of the data, this is difficult to visually inspect. On the other hand, inspection of the label assignments show a clustering of certain class labels, especially ``fire'' and ``right'', and to a lesser extent, ``left'', while ``no-op'' remains scattered throughout. The visualization of the class labels reveals a second problem: although the dataset has been rebalanced, certain classes are clearly preferred in the neuron labeling.

Test classification results for a number of hyper-parameters are gathered in Table \ref{tab:breakout_two_level_results}. For each setting of hyper-parameters, 5 experiments are run with different random seeds, and we report the average accuracy along with a single standard deviation. Notice that, for $p_\textrm{low} = 0.0$, the test results are the same across values of $c_\textrm{start}$, since the network spends no training time in the low inhibition phase. The general trend is that test results tend to increase with $n_\textrm{neurons}$, and decrease with $p_\textrm{low}$, although, in some cases, $p_\textrm{low} = 1.0$ outperforms $p_\textrm{low} \in (0, 1)$. The \textit{n-gram} classification method tends to perform better than the \textit{proportion weighting} method, which tends to outperform the \textit{all activity} method.

\begin{table}[H]
    \centering
    \footnotesize
    \begin{tabular}{||c|c|c|c|c|c||}
        \toprule
        $n_\textrm{neurons}$ & $c_\textrm{start}$ & $p_\textrm{low}$ & \textit{all activity} & \textit{proportion weighting} & \textit{n-gram} \\
\midrule
100 & 0.5 & 0.00 &  55.78 $\pm$ 0.23 &  56.38 $\pm$ 0.05 &  56.47 $\pm$ 0.80 \\
    &     & 0.25 &  48.90 $\pm$ 1.73 &  48.85 $\pm$ 1.73 &  50.22 $\pm$ 0.35 \\
    &     & 0.50 &  44.86 $\pm$ 1.84 &  44.78 $\pm$ 1.84 &  48.38 $\pm$ 1.76 \\
    &     & 1.00 &  36.41 $\pm$ 3.01 &  36.51 $\pm$ 3.19 &  37.01 $\pm$ 2.57 \\
    & 1.0 & 0.00 &  55.83 $\pm$ 2.41 &  56.35 $\pm$ 2.60 &  56.22 $\pm$ 1.32 \\
    &     & 0.25 &  50.41 $\pm$ 1.02 &  50.38 $\pm$ 1.08 &  50.15 $\pm$ 0.87 \\
    &     & 0.50 &  43.21 $\pm$ 3.47 &  43.20 $\pm$ 3.47 &  47.00 $\pm$ 3.07 \\
    &     & 1.00 &  38.97 $\pm$ 2.83 &  38.85 $\pm$ 2.75 &  38.54 $\pm$ 2.31 \\
    & 2.5 & 0.00 &  55.88 $\pm$ 2.79 &  56.32 $\pm$ 3.00 &  56.19 $\pm$ 1.52 \\
    &     & 0.25 &  49.34 $\pm$ 1.38 &  49.30 $\pm$ 1.21 &  50.14 $\pm$ 0.85 \\
    &     & 0.50 &  45.19 $\pm$ 1.42 &  45.10 $\pm$ 1.45 &  46.91 $\pm$ 2.70 \\
    &     & 1.00 &  42.49 $\pm$ 3.83 &  43.08 $\pm$ 3.19 &  41.08 $\pm$ 1.99 \\
\midrule
200 & 0.5 & 0.00 &  56.24 $\pm$ 2.57 &  56.45 $\pm$ 2.95 &  58.35 $\pm$ 1.61 \\
    &     & 0.25 &  50.46 $\pm$ 0.85 &  50.40 $\pm$ 0.84 &  50.70 $\pm$ 0.87 \\
    &     & 0.50 &  49.45 $\pm$ 1.17 &  49.39 $\pm$ 1.24 &  50.13 $\pm$ 0.88 \\
    &     & 1.00 &  40.03 $\pm$ 1.81 &  39.84 $\pm$ 2.52 &  40.57 $\pm$ 1.22 \\
    & 1.0 & 0.00 &  56.24 $\pm$ 2.57 &  56.45 $\pm$ 2.95 &  58.35 $\pm$ 1.61 \\
    &     & 0.25 &  49.92 $\pm$ 1.58 &  49.95 $\pm$ 1.42 &  50.33 $\pm$ 0.74 \\
    &     & 0.50 &  50.15 $\pm$ 0.82 &  50.11 $\pm$ 0.83 &  50.30 $\pm$ 0.38 \\
    &     & 1.00 &  40.93 $\pm$ 2.90 &  41.58 $\pm$ 2.90 &  39.59 $\pm$ 1.48 \\
    & 2.5 & 0.00 &  56.24 $\pm$ 2.57 &  56.45 $\pm$ 2.95 &  58.35 $\pm$ 1.61 \\
    &     & 0.25 &  51.26 $\pm$ 1.10 &  51.36 $\pm$ 1.34 &  50.67 $\pm$ 1.68 \\
    &     & 0.50 &  49.75 $\pm$ 1.55 &  49.98 $\pm$ 1.79 &  49.60 $\pm$ 0.81 \\
    &     & 1.00 &  50.62 $\pm$ 0.93 &  51.02 $\pm$ 1.24 &  49.90 $\pm$ 0.52 \\
\midrule
300 & 0.5 & 0.00 &  57.77 $\pm$ 1.41 &  58.19 $\pm$ 1.45 &  58.36 $\pm$ 1.55 \\
    &     & 0.25 &  51.29 $\pm$ 1.17 &  51.26 $\pm$ 1.12 &  52.21 $\pm$ 1.36 \\
    &     & 0.50 &  49.81 $\pm$ 1.75 &  49.90 $\pm$ 1.56 &  50.50 $\pm$ 1.05 \\
    &     & 1.00 &  44.73 $\pm$ 3.51 &  45.03 $\pm$ 3.23 &  43.46 $\pm$ 2.33 \\
    & 1.0 & 0.00 &  57.77 $\pm$ 1.41 &  58.19 $\pm$ 1.45 &  58.36 $\pm$ 1.55 \\
    &     & 0.25 &  50.57 $\pm$ 1.53 &  50.63 $\pm$ 1.58 &  51.96 $\pm$ 0.98 \\
    &     & 0.50 &  48.53 $\pm$ 0.86 &  48.73 $\pm$ 0.85 &  50.43 $\pm$ 0.18 \\
    &     & 1.00 &  48.20 $\pm$ 5.44 &  48.44 $\pm$ 5.29 &  46.10 $\pm$ 3.66 \\
    & 2.5 & 0.00 &  57.77 $\pm$ 1.41 &  58.19 $\pm$ 1.45 &  58.36 $\pm$ 1.55 \\
    &     & 0.25 &  52.64 $\pm$ 1.36 &  52.79 $\pm$ 1.45 &  52.46 $\pm$ 1.29 \\
    &     & 0.50 &  52.49 $\pm$ 1.50 &  52.68 $\pm$ 1.67 &  51.15 $\pm$ 0.69 \\
    &     & 1.00 &  53.55 $\pm$ 0.65 &  53.74 $\pm$ 0.47 &  52.11 $\pm$ 0.54 \\
\midrule
400 & 0.5 & 0.00 &  57.13 $\pm$ 0.80 &  57.90 $\pm$ 1.17 &  \textbf{58.93 $\pm$ 0.64} \\
    &     & 0.25 &  51.90 $\pm$ 1.15 &  51.84 $\pm$ 1.18 &  52.13 $\pm$ 1.77 \\
    &     & 0.50 &  50.41 $\pm$ 0.67 &  50.58 $\pm$ 0.69 &  51.65 $\pm$ 0.70 \\
    &     & 1.00 &  43.83 $\pm$ 3.06 &  43.77 $\pm$ 2.44 &  45.30 $\pm$ 2.33 \\
    & 1.0 & 0.00 &  57.13 $\pm$ 0.80 &  57.90 $\pm$ 1.17 &  \textbf{58.93 $\pm$ 0.64} \\
    &     & 0.25 &  51.43 $\pm$ 1.23 &  51.40 $\pm$ 1.20 &  52.84 $\pm$ 2.30 \\
    &     & 0.50 &  50.59 $\pm$ 1.78 &  50.89 $\pm$ 1.86 &  50.92 $\pm$ 1.10 \\
    &     & 1.00 &  50.88 $\pm$ 1.99 &  51.26 $\pm$ 1.93 &  49.72 $\pm$ 1.57 \\
    & 2.5 & 0.00 &  57.13 $\pm$ 0.80 &  57.90 $\pm$ 1.17 &  \textbf{58.93 $\pm$ 0.64} \\
    &     & 0.25 &  52.82 $\pm$ 2.39 &  52.90 $\pm$ 2.39 &  54.47 $\pm$ 1.64 \\
    &     & 0.50 &  52.43 $\pm$ 2.64 &  52.63 $\pm$ 2.71 &  51.78 $\pm$ 1.22 \\
    &     & 1.00 &  53.13 $\pm$ 1.80 &  53.42 $\pm$ 1.68 &  52.96 $\pm$ 1.90 \\
            \bottomrule
        \end{tabular}
    \caption{Test performance results on the Atari breakout games frames classification task using the two-level inhibition modified SNN architecture. Besides the reported varied hyper-parameters, all others are kept constant. 5 separate experiments are run, and average test results are reported along with plus or minus a single standard deviation.}
    \label{tab:breakout_two_level_results}
\end{table}

\section{Conclusions and Future Work}
\label{sec:conclusions}

We have introduced a lattice map spiking neural network (LM-SNN) model with modified STDP learning rule and biological inspired decision making mechanism. The feasibility of using LM-SNN to solve actual machine learning tasks is demonstrated on two datasets: the MNIST handwritten digits image data, and a collection of snapshots of Atari Breakout computer game images gathered from a successfully trained Deep Q-Learning networks. The introduced model generalizes previous state-of-art and demonstrates several advantages, as follows:

\begin{itemize}
\item The learning algorithm in LM-SNN manifests an unsupervised learning scheme with self-organization, classification, and clustering properties, related to Kohonen SOMs. It is shown that we can control the smoothness, or fragmentation, of the produced self-organized filter maps by dynamically tuning the model parameters. This smoothing effect helps to interpret the operation of the filters, and it is especially significant for the inhibitory connection strengths and strategies, such as incremental change in inhibition level, as well as switching inhibition from low to high values. 
Moreover, these results point to mechanisms, which may play role in the spontaneous formation of receptive fields observed in biological systems.

\item
We demonstrated that the LM-SNN model can learn fast, which is very advantageous for rapid decision-making, possibly for online learning, in dynamically changing environments. Namely, LM-SNN reaches reasonable accuracy levels within a fraction of the learning time and using much less training examples required by traditional SNNs.
To achieve such performance, we utilize the timing information in the spike patterns of the activity using the \textit{$n$-gram} scheme, which is
able to utilize the pattern and the timing of first individual spikes in order to make a predictions without waiting for a fixed amount of information. 
In addition to the ability to serve as an image classifier with clustering ability, the algorithm in this paper also show the feasibility to serve as a controller to real-time task like in the Atari Breakout game.
\item
Our approach demonstrated robustness and graceful degradation in performance in the event of missing inputs. This is in stark contrast with deep neural networks, which show extreme sensitivity and catastrophic degradation of performance with changing data properties. Specifically, the performance of LM-SNN has been degrading only slightly even if about 50$\%$ of the pixels were missing, and the performance has been maintained above random choice even with 90$\%$ of the pixels missing.
\end{itemize}


Future work will utilize the clustering ability of the excitatory layer for enhancing and scaling up this model to a multi-layer network. It is anticipated that by stacking up layers, the network would be able to cluster abstract features in the first layers and specialize the clustering features in deeper layers. Continued development of neuron labeling and classification strategies may play an important role in evaluating trained SNN activity; i.e., defining what network output entails in relation to categorical membership of the input data. Moreover, one can use this model in a real-time reinforcement learning by rewarding certain pattern activity in the classifier to manipulate the desired outcome.

Using spiking neurons has significant potential long-term benefits. With the proliferation of neuromorphic hardware platforms, the implementation of SNNs can become very efficient and cheap. Scaling up this technology is more cost-effective in term of energy and computational speed compared to the traditional neural networks. Moreover, the operation of spiking neurons is one step closer to biological neurons. Understanding how to compute with spiking neuron can give us insight for better understand computation in biological systems under normal and possibly abnormal conditions.


\section*{Acknowledgements}

This work has been supported in part by Defense Advanced Research Project Agency Grant, DARPA/MTO HR0011-16-l-0006 and by National Science Foundation Grant NSF-CRCNS-DMS-13-11165. The support of Devdhar Patel in preparing the Atari benchmark data set is greatly appreciated.

\bibliographystyle{IEEEtran}
\bibliography{bibtex.bib}

\end{document}